\documentclass[conference]{IEEEtran}
\IEEEoverridecommandlockouts
\usepackage{cite}
\usepackage{amsmath,amssymb,amsfonts}
\usepackage{algorithmic}
\usepackage{graphicx}
\usepackage{textcomp}
\usepackage[dvipsnames]{xcolor}

\usepackage{algorithmic}
\usepackage{amsmath}
\usepackage{graphicx}
\usepackage{color} 
\usepackage{algorithm} 
\usepackage{subcaption}
\usepackage{multirow}
\usepackage{multicol}
\usepackage[font=footnotesize]{caption}
\usepackage{tabularx,booktabs}

\newcommand{\boldhdr}[1]{\noindent \textbf{#1.}}

\usepackage{comment}
\usepackage{ifthen}
\newcommand{\showcomments}{yes}
\newcommand\m[1]{
    \ifthenelse{\equal{\showcomments}{no}}{{\color{red} [Ming: #1]}}{\ignorespaces}
}
\newcommand\kaiqi[1]{ % remove it, change to no
    \ifthenelse{\equal{\showcomments}{no}}{{\color{ForestGreen} [Kaiqi: #1]}}{\ignorespaces}
}
\newcommand{\kai}[1]{\textcolor{black}{#1}} % keep it, just change color!
\def\BibTeX{{\rm B\kern-.05em{\sc i\kern-.025em b}\kern-.08em
    T\kern-.1667em\lower.7ex\hbox{E}\kern-.125emX}}
\begin{document}

\title{Poster: Self-Supervised Quantization-Aware Knowledge Distillation\\
}

\author{\IEEEauthorblockN{1\textsuperscript{st} Kaiqi Zhao}
\IEEEauthorblockA{\textit{School of Computing and Augmented Intelligence} \\
\textit{Arizona State University}\\
% Arizona United States \\
kzhao27@asu.edu}
\and
\IEEEauthorblockN{2\textsuperscript{nd} Ming Zhao}
\IEEEauthorblockA{\textit{School of Computing and Augmented Intelligence} \\
\textit{Arizona State University}\\
% Arizona United States \\
mingzhao@asu.edu}
}

\maketitle

\begin{abstract}

% Quantization and Knowledge Distillation (KD) are promising approaches to compress deep learning models in order to deploy them on resource-constrained edge devices.
% Recent studies employ the combination of these two techniques to further improve the performance of compressed networks. 
Quantization-aware training (QAT) achieves competitive performance and is widely used for image classification tasks in model compression. 
Existing QAT works start with a pre-trained full-precision model and perform quantization during retraining.
However, these works require supervision from the ground-truth labels whereas sufficient labeled data are infeasible in real-world environments. 
Also, they suffer from accuracy loss due to reduced precision, and no algorithm consistently achieves the best or the worst performance on every model architecture.
To address the aforementioned limitations, this paper proposes a novel Self-Supervised Quantization-Aware Knowledge
Distillation framework (SQAKD).
SQAKD unifies the forward and backward dynamics of various quantization functions, making it flexible for incorporating the various QAT works.
With the full-precision model as the teacher and the low-bit model as the student, SQAKD reframes QAT as a co-optimization problem that simultaneously minimizes the KL-Loss (i.e., the Kullback-Leibler divergence loss between the teacher’s and student’s penultimate outputs) and the discretization error (i.e., the difference between the full-precision weights/activations and their quantized counterparts).
This optimization is achieved in a self-supervised manner without labeled data.
The evaluation shows that SQAKD significantly improves the performance of various state-of-the-art QAT works (e.g., PACT, LSQ, DoReFa, and EWGS).
SQAKD establishes stronger baselines and does not require extensive labeled training data, potentially making state-of-the-art QAT research more accessible.
% We hope this will make state-of-the-art QAT research more accessible. 

% Code will be made public.

% We are the first to 1) investigate 11 KD methods in the context of Quantization-Aware Training (QAT), and 2) analyze the loss function of Knowledge Distillation (KD) in the quantization problem.
% The proposed SQAKD framework can improve the performance of a large amount of state-of-the-art quantization methods, e.g., PACT, LSQ, DoReFa, and EWGS, by a large margin.

% \begin{IEEEkeywords}
% component, formatting, style, styling, insert
% \end{IEEEkeywords}

\end{abstract}

\section{Introduction}

Quantization is one of the model compression approaches~\cite{zhao2022knowledge, zhao2023contrastive, zhao2023automatic, carreira2021model, idelbayev2021lc} to address the mismatch issue between resource-hungry DNNs and resource-constrained edge devices~\cite{chen2019exploring, zhaoknowledgenet, chen2019exploring2}.
Various quantization techniques~\cite{zhou2016dorefa, choi2018pact, esser2019learned, lee2021network} have achieved great results in creating low-bit models through Quantization-Aware Training (QAT), which starts with a pre-trained model and performs quantization during retraining.
% Existing works on QAT, e.g. DoReFa-Net~\cite{zhou2016dorefa}, PACT~\cite{choi2018pact}, LSQ~\cite{esser2019learned}, APoT~\cite{li2019additive}, DSQ~\cite{gong2019differentiable}, QIL~\cite{jung2019learning} and EWGS~\cite{lee2021network}, 
% focus on designing the proper format of the forward and backward of the quantization function.
% have primarily directed their efforts toward refining the intricacies of forward and backward formulation of the quantization function.
\m{why do they suffer if the results are great? \kai{Changed the sentence}}
% However, most of them suffer from accuracy loss due to quantization~\cite{boo2021stochastic, li2021mqbench} and no algorithm consistently achieves the best or the worst performance on every model architecture (e.g., VGG, ResNet, MobileNet, etc.)~\cite{li2021mqbench}.
However, most of them \kai{result in some degree of accuracy loss} due to reduced precision~\cite{boo2021stochastic} and no algorithm consistently achieves the best or the worst performance on every model architecture (e.g., VGG, ResNet, MobileNet, etc.)~\cite{li2021mqbench}.
Also, the various QAT works are motivated by different intuitions and lack a commonly agreed theory, which makes it challenging to generalize.
Moreover, all the QAT works assume labeled training data are always available \kai{while labeling the data can be time-consuming and sometimes even infeasible, particularly in specialized domains or for specific tasks}\m{explain why this is a bad assumption \kai{added}}.

To address the aforementioned limitations and improve the performance of the state-of-the-art (SOTA) QAT for model compression, we propose a simple yet effective framework --- Self-Supervised Quantization-Aware Knowledge Distillation (SQAKD). 
SQAKD first unifies the forward and backward dynamics of various quantization functions and shapes QAT as minimizing the discretization error between original weights/activations and their quantized counterparts. \kaiqi{Think more and modify methodology}
Then, SQAKD lets the full-precision and low-bit models as the teacher and student, respectively,
and excludes CE-Loss (i.e., cross-entropy loss with labels) and keeps only the KL-Loss (i.e., the Kullback-Leibler divergence loss between the teacher’s and student’s penultimate outputs), reframing QAT as a co-optimization problem that simultaneously minimizes the KL-Loss and the discretization error without supervision from labels.

% Compared to existing QAT methods and the combined methods integrating both KD and QAT, the proposed SQAKD has several advantages. 
Compared to existing QAT methods, SQAKD has several advantages. 
First, SQAKD is flexible for incorporating the various QAT works \kai{since it unifies the optimization of their forward and backward dynamics.} \m{what's a quantizer? how does SQAKD incoprates them? \kai{added}}
% and is effective in improving their performance \kai{by incorporating guidance from the full-precision model to inform gradient updates.} \m{how? \kai{added}}
% Second, SQAKD is hyperparameter-free, since it only uses the KL-Loss as the training loss and does not require the weight adjustment of multiple loss terms that appear in previous methods.
Second, SQAKD operates in a self-supervised manner without labeled data, supporting a more extensive scope of applications in practical scenarios.
Finally, SQAKD improves the various SOTA QAT works significantly in both convergence speed and accuracy.

\section{Methodology}

\begin{figure}[t]
    \centering
    \includegraphics[width=\linewidth]{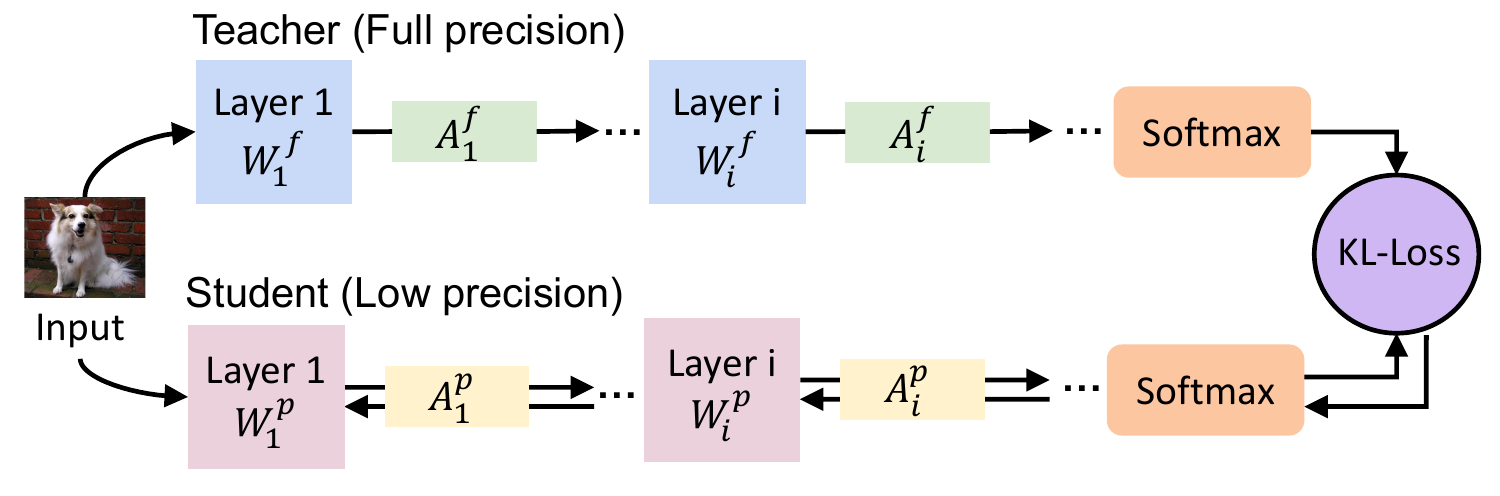}
    \vspace{-15pt}
    \caption{\scriptsize{Workflow of SQAKD.}}
    \label{fig:workflow}
    \vspace{-10pt}
\end{figure}

% \subsection{Framework Overview}

% Figure~\ref{fig:workflow} \kaiqi{working on} illustrates the workflow of the proposed framework --- Self-Supervised Quantization-Aware Knowledge Distillation (SQAKD). 
Figure~\ref{fig:workflow} \kaiqi{working on} illustrates the workflow of the proposed SQAKD. 
%
% We first formulate a generalized formulation for both forward and backward propagation in the context of quantization,
% then redefine the quantization as a Knowledge distillation (KD)-based optimization problem.

% \subsubsection{Forward Propagation}
\smallskip
\boldhdr{Forward Propagation}
Let $Quant(\cdot)$ denote a uniform quantizer that converts a full-precision input $x$ to a quantized output $x_q = Quant(x)$.
$x$ can be the activations or weights of the network.
First, the quantizer $Quant(\cdot)$ applies a clipping function $Clip(\cdot)$ to normalize and clip the input $x$ to smaller range, producing a full-precision latent presentation: $x_c = Clip(x, \{p_i\}_{i=1}^{i=K_c}, v, m)$,
% \begin{equation}\label{tab:clip_function}
%     x_c = Clip(x, \{p_i\}_{i=1}^{i=K_c}, v, m),
% \end{equation}
where $v$ and $m$ are the lower and upper bound of the range, respectively,
$\{p_i\}_{i=1}^{i=K_c}$ denotes the set of trainable parameters needed for quantization,
and $K_c$ denotes the number of parameters.
%
% Different quantizers have different schemes for $Clip(\cdot)$.
% For example, in PACT~\cite{choi2018pact}, the lower bound $v$ is set to 0 and the upper bound $m$ is a trainable parameter optimized during training. 
% The quantizer only requires one parameter, i.e., $\{p_1|p_1=m, K_c=1\}$,
% and the clipping function is described as: $x_c = Clip(x, \{p_1|p_1=m\}, 0, m) = 0.5(|x|-|x-m|+m)$.
% %
% In EWGS~\cite{lee2021network}, $v$ and $m$ are set to 0 and 1, respectively, and every quantized layer use separate parameters (i.e., $p_1$ and $p_2$) for the quantization intervals: $x_c = Clip(x, \{p_1, p_2\}, 0, 1) = clip(\frac{x-p_1}{p_1-p_2}, 0, 1)$.
% % In LSQ~\cite{}, for $b$-bit quantization, if $x$ is activations, $l=0$ and $u=2^b-1$; and if $x$ is weights, $l=2^{b-1}$ and $u=2^{b-1}-1$.  
%
Then, the quantizer $Quant(\cdot)$ converts the clipped value $x_c$ to a discrete quantization point $x_q$ using the function $R(\cdot)$ that contains a round function: $x_q = R(x_c, b, \{q_i\}_{i=1}^{i=K_r})$,
% \begin{equation}\label{equ:round_function}
%     x_q = R(x_c, b, \{q_i\}_{i=1}^{i=K_r})
% \end{equation}
where $b$ is the bit width and $\{q_i\}_{i=1}^{i=K_r}$ denotes the set of trainable parameters.
%
% Note that $\{q_i\}_{i=1}^{i=K_r}$ is not necessary for some quantizers, like EWGS~\cite{lee2021network} and DoReFa~\cite{zhou2016dorefa}.
% For example, in EWGS, if activations are the input, $x_q = R(x_c, b) = \frac{round((2^b-1) \cdot x_c)}{2^b-1}$; and if weights are the input, $x_q = R(x_c, b) = 2(\frac{round((2^b-1) \cdot x_c)}{2^b-1} - 0.5)$.
% %
% In some quantizers, like PACT~\cite{choi2018pact} and LSQ~\cite{esser2019learned}, the trainable parameters in the function $R(\cdot)$ are the same as those in the clipping function $Clip(\cdot)$, i.e., $\{q_i|q_i=p_i\}_{i=1}^{i=K_r}$ and $K_r=K_c$.
% For example, in PACT, $x_q = R(x_c, b, \{q_1|q_1=p_1\}) = round(x_c \cdot \frac{2^b-1}{q_1}) \cdot \frac{q_1}{2^b-1}$.
%
% In summary, the quantization function 
Thus, $Quant(\cdot)$ can be described as: $x_q = Quant(x, \alpha, b, v, m)$,
% \begin{equation}
%     % x_q = Quant(x, \{p_i\}_{i=1}^{i=K_c}, \{q_i\}_{i=1}^{i=K_r}, b, v, m).
%     x_q = Quant(x, \alpha, b, v, m),
% \end{equation}
here we use $\alpha$ as a shorthand for the set of all the parameters in the functions $R(\cdot)$ and $Clip(\cdot)$: $\alpha = \{ \{p_i\}_{i=1}^{i=K_c}, \{q_i\}_{i=1}^{i=K_r} \}$.
% Specifically, for weights $W$ and activations $A$ as the input, their corresponding quantization function produces the outputs $W_q = Quant_W(W, \alpha_W, b_W, v_W, m_W)$ and $A_q = Quant_A(A, \alpha_A, b_A, v_A, m_A)$, respectively.
% , as follows: $W_q = Quant_W(W, \alpha_W, b_W, v_W, m_W)$, and $A_q = Quant_A(A, \alpha_A, b_A, v_A, m_A)$.
% \begin{equation}
% \begin{split}
%     W_q = Quant_W(W, \alpha_W, b_W, v_W, m_W) \\
%     A_q = Quant_A(A, \alpha_A, b_A, v_A, m_A).  
% \end{split}
% \end{equation}

% Clipping: Confine weights/activations to a smaller range [min, max]
% Rounding: Map the clipped value to its nearest quantization point with a scaling factor

\smallskip
\boldhdr{Backward Propagation}
% Directly training a quantized network using the back-propagation is impossible since the quantization function $Q(\cdot)$ is non-differentiable.
% This issue arises due to the round function in $R(\cdot)$, which produces near-zero derivatives almost everywhere.
% To solve this problem, most of the quantization methods~\cite{choi2018pact,  esser2019learned, zhou2016dorefa, gong2019differentiable, jung2019learning} use Straight-Through Estimator (STE)~\cite{bengio2013estimating} to approximate the gradients: $\frac{\partial L}{\partial x_c} = \frac{\partial L}{\partial x_q}$.
% %
% Recently, an alternative back-propagation method to the widely used STE is proposed in EWGS~\cite{lee2021network}.
% We find that the success of EWGS is mainly due to its integration of the discretization error, defined as $x_c-x_q$, within the gradient approximation process (see Section~\ref{sec:ablation} for evaluation results).
% Based on this finding, in this work, we provide a generalization of the two aforementioned branches of backward-propagation techniques using the following equation: 
Different from prevalent quantization works using Straight-Through Estimator (STE)~\cite{bengio2013estimating} in backpropagation, we propose a novel formulation of backpropagation by integrating the discretization error ($x_c - x_q$): 
% $\frac{\partial L}{\partial x_c} = \frac{\partial L}{\partial x_q} + \mu \cdot (x_c - x_q)$,
\vspace{-3pt}
\begin{equation}
    \small
    \frac{\partial L}{\partial x_c} = \frac{\partial L}{\partial x_q} + \mu \cdot (x_c - x_q),
\vspace{-3pt}
\end{equation}
where $\mu$ is a non-negative value.
% When $\mu$ equals to zero, it is STE; and when $\mu$ equals to $\delta \cdot sign(\frac{\partial L}{\partial x_q}) \cdot \frac{\partial L}{\partial x_q}$, it is EWGS.
STE is represented by setting $\mu$ to zero, and $\mu$ can also be updated by other schemes, like Curriculum Learning driven strategy~\cite{bengio2009curriculum, zhao2022knowledge}.

% , while EWGS is obtained when $\mu$ is equal to the multiply of $\delta$, $sign(\frac{\partial L}{\partial x_q})$, and $\frac{\partial L}{\partial x_q}$.
% Note that $\mu$ can also be updated by other schemes, like Curriculum Learning driven strategy~\cite{bengio2009curriculum, zhao2022knowledge}.
% \kaiqi{reference format}

% Thus, the quantization-aware training problem is defined as finding the optimal parameters $\alpha_W$ and $\alpha_A$ to minimize the squared error between original weights/activations and their quantized counterparts, and also finding the unquantized model weights $W$ to minimize the discrepancy between the model predictions with the true labels.
% The aim is to achieve accurate quantization while ensuring that the quantized model retains its predictive capability.
% The objective function can be defined as follows:

% \subsection{Optimization via Self-Supervised Knowledge Distillation}
% \label{sec:SQAKD}
% \subsubsection{
\smallskip
\boldhdr{Optimization Objective}
% Knowledge distillation (KD)~\cite{hinton2015distilling} has been widely shown to perform well for improving the performance of a small network (``student'') by transferring the information from a large network (``teacher'').
% However, the effectiveness of KD in achieving the optimal solution to the quantization-aware training problem remains unexplored.
%
To apply Knowledge Distillation (KD)~\cite{hinton2015distilling} into quantization, we let a pre-trained full-precision network act as the teacher, guiding a low-bit student with the same architecture.
% Then, to enable the low-bit student to benefit from the guidance of the teacher's full-precision gradients, the optimization process involves minimizing the distillation loss (termed ``KL-Loss''), which captures the difference in output distributions between the teacher and student, and the cross-entropy loss (term ``CE-Loss''), which measures the discrepancy between the student's predictions and the ground-truth labels.
% KD~\cite{hinton2015distilling} defines the distillation loss as the KL divergence between teacher and student soft logits, and the overall training loss as a linear combination of the cross-entropy loss with labels (term ``CE-Loss'') and the distillation loss (termed ``KL-Loss''), controlled by a hyper-parameter $\lambda$: $L = (1-\lambda) L_{CE} + \lambda L_{KL}$.
 KD~\cite{hinton2015distilling} defines the training loss as a linear combination of the cross-entropy loss (term ``CE-Loss'') with labels and the KL divergence (termed ``KL-Loss'') between the teacher's and student's soft logits, controlled by a hyper-parameter $\lambda$: $L = (1-\lambda) L_{CE} + \lambda L_{KL}$.
% where $\lambda$ is a hyper-parameter that controls the weight of the loss terms.

However, we find out that CE-Loss does not cooperate effectively with KL-Loss, and their combination may potentially degrade the network performance. 
Solely minimizing KL-Loss is sufficient for achieving optimal gradient updates in the quantized network.
% Motivated by the observations from Section~\ref{sec:analysis_of_kd_on_quantization}, 
So we drop the CE-Loss and keep only the KL-Loss in SQAKD.
The optimization objection is defined as:
\vspace{-3pt}
\begin{equation}\label{equ:optimization_SQAKD}
    \small
    \begin{split}
    \underset {W^S_f, \alpha_W, \alpha_A} {min} 
    % &\frac{\mu_W^S ||W^S - W_q^S||^2}{2} + \frac{\mu_A^S ||A^S - A_q^S||^2}{2} \\
    &KL (S(h^T/\rho) || S(h^S/\rho)) \\
    s.t. \quad &W_q^S = Quant_W(W^S_f, \alpha_W, b_W, v_W, m_W) \\
    &A_q^S = Quant_A(A^S_f, \alpha_A, b_A, v_A, m_A),
    \end{split} 
\vspace{-5pt}
\end{equation}
where $\rho$ is the temperature, which makes distribution softer for using the dark knowledge,
$Y$ is the ground-truth labels, 
$h^T$ and $h^S$ are the penultimate layer outputs of the teacher and student, respectively.
$Quant_W(\cdot)$ and $Quant_A(\cdot)$ are the quantization function for the student's weights and activations,
and $W^S_f$/$A^S_f$ and $W^S_q$/$A^S_q$ are the student's full-precision weights/activations and quantized weights/activations.

% Specifically, for weights $W$ and activations $A$ as the input, their corresponding quantization function produces the outputs $W_q = Quant_W(W, \alpha_W, b_W, v_W, m_W)$ and $A_q = Quant_A(A, \alpha_A, b_A, v_A, m_A)$, respectively.

%
% During training, the teacher's weights are frozen and it performs the forward propagation only.
% In the forward pass, the student's parameters are quantized, while the corresponding full-precision values are preserved internally. 
% During the backward propagation, the gradients are applied to the student's preserved full-precision values.
% After convergence, the student retains its full-precision weights and the parameters utilized in the quantization function. 
% The student's quantized weights are obtained by applying the full-precision values to the quantization function.

\section{Evaluation}

% \usepackage{booktabs}
\begin{table}[t]
% \vspace{-5pt}
\scriptsize
\centering
\caption{Top-1 test accuracy (\%) on CIFAR-10 and CIFAR-100.}
\vspace{-6pt}
% ``FP'' denotes the original full-precision model, and ``W$*$A$\times$'' denotes that the weights and activations are quantized into $*$bit and $\times$bit, respectively. \textcolor{blue}{Blue} numbers inside the brackets denote the increase compared to the quantization method alone. SQAKD (EWGS) denotes models when the EWGS~\cite{lee2021network} quantizer is used in the proposed SQAKD framework.}
\label{tab:cifar_improvement_on_quantization}
\setlength{\tabcolsep}{1.4pt}
\begin{tabularx}{0.45\textwidth}{@{}cl|ccc|ccc@{}}
% \begin{tabularx}{0.45\textwidth}{@{}clcccccc@{}}
\toprule
\multirow{4}{*}{CIFAR-10}  & Model                 & \multicolumn{3}{c}{\begin{tabular}[c]{@{}c@{}}VGG-8\\ (FP: 91.27)\end{tabular}}                                                                                                                        & \multicolumn{3}{c}{\begin{tabular}[c]{@{}c@{}}ResNet-20\\ (FP: 92.58)\end{tabular}}                                                                                                                    \\ \cmidrule(l){2-8} 
                           & Bit-width                   & W1A1                                                             & W2A2                                                             & W4A4                                                             & W1A1                                                             & W2A2                                                             & W4A4                                                             \\
                           & EWGS~\cite{lee2021network}                  & 87.77                                                            & 90.84                                                            & 90.95                                                            & 86.42                                                            & 91.41                                                            & 92.40                                                            \\
                           & \textbf{SQAKD (EWGS)} & \textbf{\begin{tabular}[c]{@{}c@{}}89.05\\ \textcolor{blue}{(+1.28)}\end{tabular}} & \textbf{\begin{tabular}[c]{@{}c@{}}91.55\\ \textcolor{blue}{(+0.71)}\end{tabular}} & \textbf{\begin{tabular}[c]{@{}c@{}}91.31\\ \textcolor{blue}{(+0.36)}\end{tabular}} & \textbf{\begin{tabular}[c]{@{}c@{}}86.47\\ \textcolor{blue}{(+0.05)}\end{tabular}} & \textbf{\begin{tabular}[c]{@{}c@{}}91.80\\ \textcolor{blue}{(+0.39)}\end{tabular}} & \textbf{\begin{tabular}[c]{@{}c@{}}92.59\\ \textcolor{blue}{(+0.19)}\end{tabular}} \\ \midrule
\multirow{4}{*}{CIFAR-100} & Model                 & \multicolumn{3}{c}{\begin{tabular}[c]{@{}c@{}}VGG-13\\ (FP: 76.36)\end{tabular}}                                                                                                                       & \multicolumn{3}{c}{\begin{tabular}[c]{@{}c@{}}ResNet-32\\ (FP: 71.33)\end{tabular}}                                                                                                                    \\ \cmidrule(l){2-8} 
                           & Bit-width                   & W1A1                                                             & W2A2                                                             & W4A4                                                             & W1A1                                                             & W2A2                                                             & W4A4                                                             \\
                           & EWGS~\cite{lee2021network}                  & 65.55                                                            & 73.31                                                            & 73.41                                                            & 59.25                                                            & 69.37                                                            & 70.50                                                            \\
                           & \textbf{SQAKD (EWGS)} & \textbf{\begin{tabular}[c]{@{}c@{}}68.56\\ \textcolor{blue}{(+3.01)}\end{tabular}} & \textbf{\begin{tabular}[c]{@{}c@{}}74.65\\ \textcolor{blue}{(+1.34)}\end{tabular}} & \textbf{\begin{tabular}[c]{@{}c@{}}74.67\\ \textcolor{blue}{(+1.26)}\end{tabular}} & \textbf{\begin{tabular}[c]{@{}c@{}}59.41\\ \textcolor{blue}{(+0.16)}\end{tabular}} & \textbf{\begin{tabular}[c]{@{}c@{}}69.99\\ \textcolor{blue}{(+0.62)}\end{tabular}} & \textbf{\begin{tabular}[c]{@{}c@{}}71.65\\ \textcolor{blue}{(+1.15)}\end{tabular}} \\ \bottomrule
\end{tabularx}
% \vspace{-5pt}
\end{table}

\begin{table}[t]
% \vspace{-5pt}
\scriptsize
\centering
\caption{Top-1 test accuracy (\%) of ResNet and VGG on Tiny-ImageNet.}
% \caption{Top-1 test accuracy (\%) of ResNet-18 and VGG-11 on Tiny-ImageNet.}
% \caption{Top-1 test accuracy (\%) on Tiny-ImageNet.}
\vspace{-6pt}
% ``FP'' denotes the original full-precision model, and ``W$*$A$\times$'' denotes that the weights and activations are quantized into $*$bit and $\times$bit, respectively. \textcolor{blue}{Blue} numbers inside the brackets denote the increase compared to the quantization method alone. SQAKD (PACT), SQAKD (LSQ), SQAKD (DoReFa) denote models when PACT~\cite{choi2018pact}, LSQ~\cite{esser2019learned}, and DoReFa~\cite{zhou2016dorefa} quantizers are used in our framework, respectively.}
\label{tab:tiny_imagenet_improvement_on_quantization_large_models}
\setlength{\tabcolsep}{4.0pt}
% \begin{tabularx}{0.45\textwidth}{@{}l|lll|lll@{}}
\begin{tabularx}{0.45\textwidth}{@{}l|lll|lll@{}}
\toprule
Model                   & \multicolumn{3}{c}{\begin{tabular}[c]{@{}c@{}}ResNet-18\\ (FP: 65.59)\end{tabular}}                                                                                                                   & \multicolumn{3}{c}{\begin{tabular}[c]{@{}c@{}}VGG-11\\ (FP: 59.47)\end{tabular}}                                                                                                                       \\ \midrule
Bit-width               & W3A3                                                             & W4A4                                                             & W8A8                                                             & W3A3                                                             & W4A4                                                             & W8A8                                                             \\
PACT~\cite{choi2018pact}                    & 58.09                                                            & 61.06                                                            & 64.91                                                            & 52.94                                                            & 57.10                                                            & 58.08                                                            \\
\textbf{SQAKD (PACT)}   & \textbf{\begin{tabular}[c]{@{}l@{}}61.34\\ \textcolor{blue}{(+3.25)}\end{tabular}} & \textbf{\begin{tabular}[c]{@{}l@{}}61.47\\ \textcolor{blue}{(+0.41)}\end{tabular}} & \textbf{\begin{tabular}[c]{@{}l@{}}65.78\\ \textcolor{blue}{(+0.87)}\end{tabular}} & \textbf{\begin{tabular}[c]{@{}l@{}}57.25\\ \textcolor{blue}{(+4.31)}\end{tabular}} & \textbf{\begin{tabular}[c]{@{}l@{}}59.05\\ \textcolor{blue}{(+1.95)}\end{tabular}} & \textbf{\begin{tabular}[c]{@{}l@{}}59.44\\ \textcolor{blue}{(+1.36)}\end{tabular}} \\ \midrule
LSQ~\cite{esser2019learned}                     & 61.99                                                            & 64.10                                                            & 65.08                                                            & 58.39                                                            & 59.14                                                            & 59.25                                                            \\
\textbf{SQAKD (LSQ)}    & \textbf{\begin{tabular}[c]{@{}l@{}}65.21\\ \textcolor{blue}{(+3.22)}\end{tabular}} & \textbf{\begin{tabular}[c]{@{}l@{}}65.34\\ \textcolor{blue}{(+1.24)}\end{tabular}} & \textbf{\begin{tabular}[c]{@{}l@{}}65.96\\ \textcolor{blue}{(+0.88)}\end{tabular}} & \textbf{\begin{tabular}[c]{@{}l@{}}58.43\\ \textcolor{blue}{(+0.04)}\end{tabular}} & \textbf{\begin{tabular}[c]{@{}l@{}}59.19\\ \textcolor{blue}{(+0.05)}\end{tabular}} & \textbf{\begin{tabular}[c]{@{}l@{}}59.42\\ \textcolor{blue}{(+0.17)}\end{tabular}} \\ \midrule
DoReFa~\cite{zhou2016dorefa}                  & 61.94                                                            & 62.72                                                            & 63.23                                                            & 56.72                                                            & 57.28                                                            & 57.54                                                            \\
\textbf{SQAKD (DoReFa)} & \textbf{\begin{tabular}[c]{@{}l@{}}64.10\\ \textcolor{blue}{(+2.16)}\end{tabular}} & \textbf{\begin{tabular}[c]{@{}l@{}}64.56\\ \textcolor{blue}{(+1.84)}\end{tabular}} & \textbf{\begin{tabular}[c]{@{}l@{}}64.88\\ \textcolor{blue}{(+1.65)}\end{tabular}} & \textbf{\begin{tabular}[c]{@{}l@{}}57.02\\ \textcolor{blue}{(+0.3)}\end{tabular}}  & \textbf{\begin{tabular}[c]{@{}l@{}}58.93\\ \textcolor{blue}{(+1.65)}\end{tabular}} & \textbf{\begin{tabular}[c]{@{}l@{}}58.91\\ \textcolor{blue}{(+1.37)}\end{tabular}} \\ \bottomrule
\end{tabularx}
% \vspace{-15pt}
% \vspace{-5pt}
\end{table}

% \usepackage{booktabs}
% \usepackage{multirow}
\begin{table}[t]
% \vspace{-5pt}
\scriptsize
\centering
\caption{Top-1 and top-5 test accuracy (\%) of MobileNet-V2, ShuffleNet-V2, and SqueezeNet on Tiny-ImageNet.}
\vspace{-6pt}
% ``FP'' denotes the original full-precision model, and ``W$*$A$\times$'' denotes that the weights and activations are quantized into $*$bit and $\times$bit, respectively. \textcolor{blue}{Blue} numbers inside the brackets denote the increase compared to the quantization method alone. SQAKD (PACT), SQAKD (LSQ), SQAKD (DoReFa) denote models when PACT~\cite{choi2018pact}, LSQ~\cite{esser2019learned}, and DoReFa~\cite{zhou2016dorefa} quantizers are used in our framework, respectively.}
\label{tab:tiny_imagenet_improvement_on_quantization_small_models}
\setlength{\tabcolsep}{3.0pt}
\begin{tabularx}{0.45\textwidth}{@{}lllll@{}}
\toprule
Model                          & Bit-width             & Method                  & Top-1 Acc.               & Top-5 Acc.              \\ \midrule
\multirow{7}{*}{MobileNet-V2}  & FP                    & -                       & 58.07                    & 80.97                   \\
                               & \multirow{2}{*}{W3A3} & PACT~\cite{choi2018pact}                    & 47.77                    & 73.44                   \\
                               &                       & \textbf{SQAKD (PACT)}   & \textbf{52.73 \textcolor{blue}{(+4.96)}}   & \textbf{77.68 \textcolor{blue}{(+4.24)}}  \\
                               & \multirow{2}{*}{W4A4} & PACT~\cite{choi2018pact}                    & 50.33                    & 75.08                   \\
                               &                       & \textbf{SQAKD (PACT)}   & \textbf{57.14 \textcolor{blue}{(+6.81)}}   & \textbf{80.61 \textcolor{blue}{(+5.53)}}  \\
                               & \multirow{2}{*}{W8A8} & DoReFa~\cite{zhou2016dorefa}                  & 56.26                    & 79.64                   \\
                               &                       & \textbf{SQAKD (DoReFa)} & \textbf{58.13 \textcolor{blue}{(+1.87)}}   & \textbf{81.3 \textcolor{blue}{(+1.66)}}   \\ \midrule
\multirow{5}{*}{ShuffleNet-V2} & FP                    & -                       & 49.91                    & 76.05                   \\
                               & \multirow{2}{*}{W4A4} & PACT~\cite{choi2018pact}                    & 27.09                    & 52.54                   \\
                               &                       & \textbf{SQAKD (PACT)}   & \textbf{41.11  \textcolor{blue}{(+14.02)}} & \textbf{68.4 \textcolor{blue}{(+15.86)}}  \\
                               & \multirow{2}{*}{W8A8} & DoReFa~\cite{zhou2016dorefa}                  & 45.96                    & 71.93                   \\
                               &                       & \textbf{SQAKD (DoReFa)} & \textbf{47.33 \textcolor{blue}{(+1.37)}}   & \textbf{73.85 \textcolor{blue}{(+1.92)}}  \\ \midrule
\multirow{5}{*}{SqueezeNet1\_0}    & FP                    & -                       & 51.49                    & 76.02                   \\
                               & \multirow{2}{*}{W4A4} & LSQ~\cite{esser2019learned}                     & 35.37                    & 62.75                   \\
                               &                       & \textbf{SQAKD (LSQ)}    & \textbf{47.40  \textcolor{blue}{(+12.03)}} & \textbf{73.18 \textcolor{blue}{(+10.43)}} \\
                               & \multirow{2}{*}{W8A8} & DoReFa~\cite{zhou2016dorefa}                  & 42.66                    & 69.25                   \\
                               &                       & \textbf{SQAKD (DoReFa)} & \textbf{46.62  \textcolor{blue}{(+3.96)}}  & \textbf{73.02 \textcolor{blue}{(+3.77)}}  \\ \bottomrule
\end{tabularx}
\vspace{-5pt}
\end{table}

\begin{figure}[t]
  \centering
  \captionsetup[subfigure]{justification=centering}
  \begin{subfigure}{.305\linewidth}
    \centering
    \includegraphics[width=2.8cm]{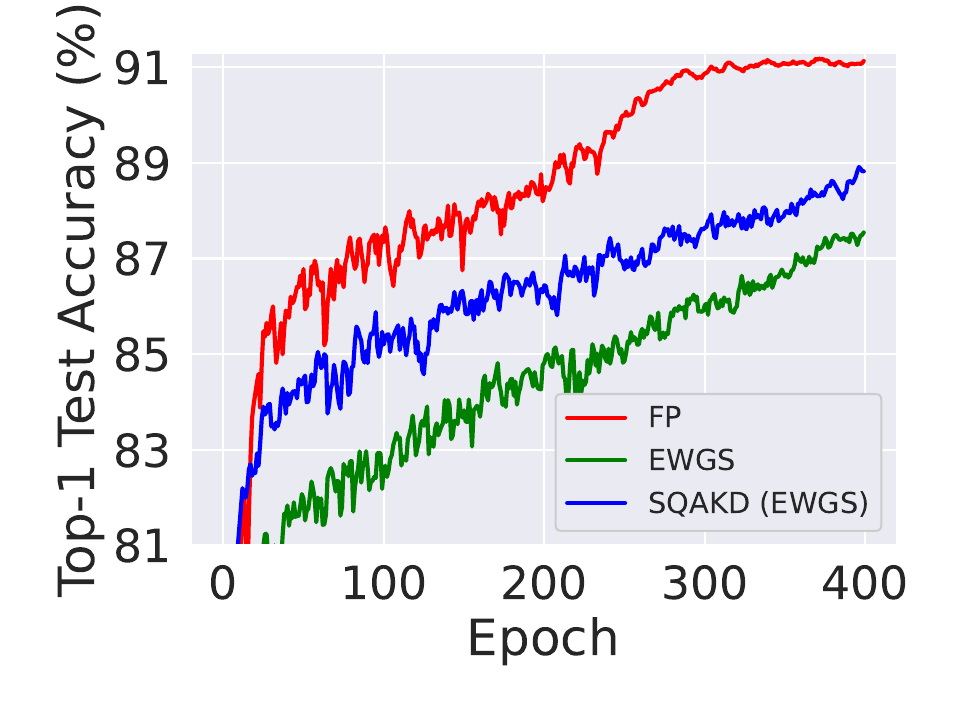}
    \vspace{-15pt}
    \caption{\scriptsize{VGG-8 \\ (W1A1, CIFAR-10)}}
    % \caption{\footnotesize{VGG-8 \\(W1A1, CIFAR-10)}}
    \label{fig:cifar10_vgg8_improvement_on_quantizatio}
  \end{subfigure}
  \begin{subfigure}{.305\linewidth}
    \centering
    \includegraphics[width=2.8cm]{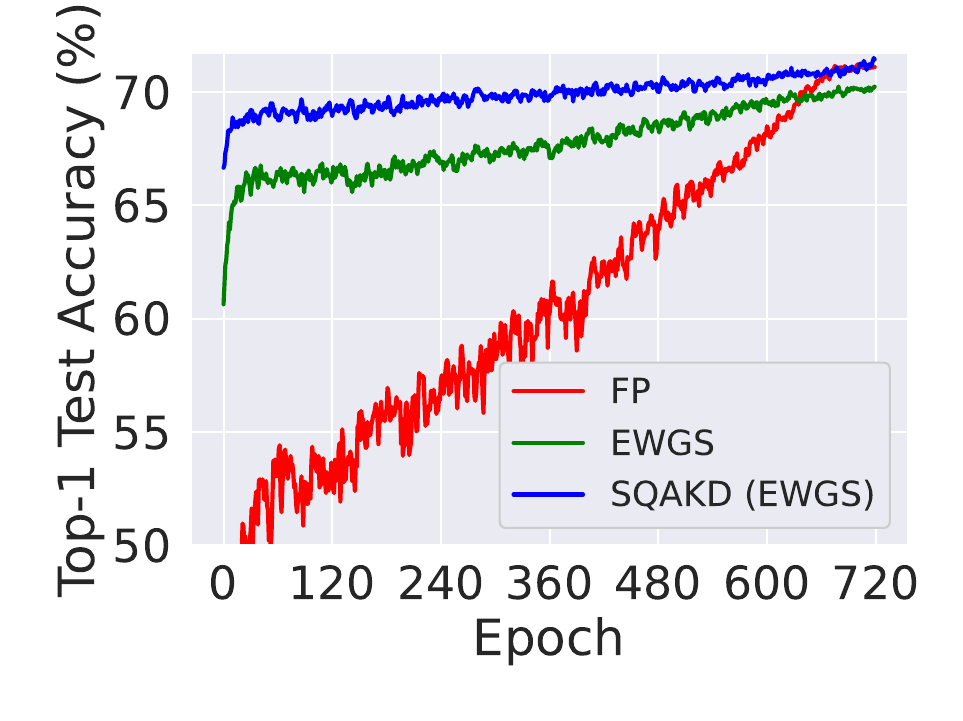}
    \vspace{-15pt}
    \caption{\scriptsize{ResNet-32 \\ (W4A4, CIFAR-100)}}
    % \caption{\footnotesize{ResNet-32 \\(W4A4, CIFAR-100)}}
  \end{subfigure}
  \begin{subfigure}{.305\linewidth}
    \centering
    \includegraphics[width=2.8cm]{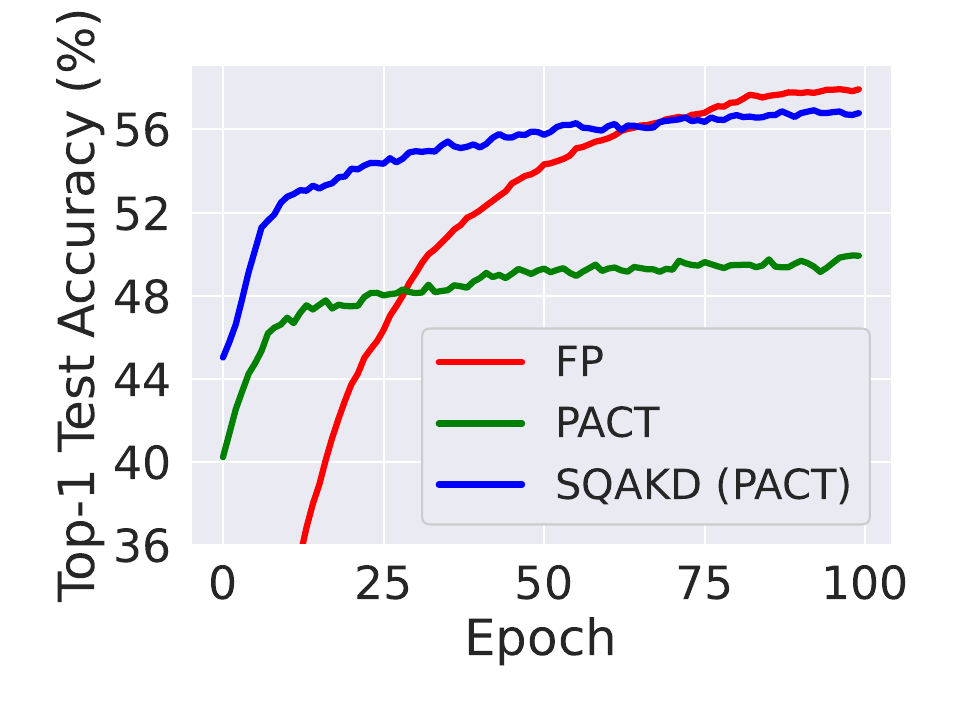}
    \vspace{-15pt}
    \caption{\scriptsize{MobileNet-V2 (W4A4, Tiny-ImageNet)}}
    % \caption{\footnotesize{ResNet-20 (W2A2, CIFAR-10)}}
  \end{subfigure}
  \vspace{-4pt}
  \caption{Top-1 test accuracy evolution of full-precision models (FP), and quantized models using EWES and EWGS+SQAKD, in each epoch during training on CIFAR-10 and CIFAR-100. ``W$*$A$\times$'' denotes that the weights and activations are quantized into $*$bit and $\times$bit, respectively.}
  % (a) VGG-8 (W1A1) and (b) ResNet-20 (W2A2) with CIFAR-10, and (c) VGG-13 (W1A1) and (d) ResNet-32 (W4A4) with CIFAR-100.
  \label{fig:improvement_on_quantization}
    \vspace{-10pt}
\end{figure}

\smallskip
\boldhdr{Experiment Setup}
%
% \subsubsection{Models and Datasets}
We conduct an extensive evaluation on diverse models (ResNet~\cite{he2016identity}, VGG~\cite{eigen2015predicting}, MobileNet~\cite{sandler2018mobilenetv2}, ShuffleNet~\cite{zhang2018shufflenet}, SqueezeNet~\cite{iandola2016squeezenet}, and AlexNet~\cite{krizhevsky2012imagenet}) and various datasets, including CIFAR-10~\cite{krizhevsky2009learning}, CIFAR-100~\cite{krizhevsky2009learning}, and Tiny-ImageNet~\cite{le2015tiny}.
We incorporate state-of-the-art (SOTA) quantization methods (PACT~\cite{choi2018pact}, LSQ~\cite{esser2019learned}, DoReFa~\cite{zhou2016dorefa}, and EWGS~\cite{lee2021network}) into SQAKD, and compare to their original results to show the improvement made by SQAKD.
%
% In Tables~\ref{tab:cifar_improvement_on_quantization}, \ref{tab:tiny_imagenet_improvement_on_quantization_large_models}, \ref{tab:tiny_imagenet_improvement_on_quantization_small_models}, 
% ``FP'' denotes the original full-precision model, and ``W$*$A$\times$'' denotes that the weights and activations are quantized into $*$bit and $\times$bit, respectively. \textcolor{blue}{Blue} numbers inside the brackets denote the increase compared to the quantization method alone. 

% SQAKD ($##$) denotes models when the quantizer $##$ is used in the proposed SQAKD framework.

% We implement SQAKD on PyTorch version 1.10.0 and Python version 3.9.7 and conduct experiments on four Nvidia RTX 2080 GPUs.
% %
% On CIFAR-10, we use Adam optimizer to train VGG-8 and ResNet-20 for 400 and 1200 epochs, respectively. 
% Weight decay is set to 1e-4. 
% The learning is set to 1e-3 initially and decays to 0.0 using a cosine annealing schedule. 
% %
% On CIFAR-100, the Adam optimizer is used to train the models for 720 epochs. 
% Weight decay is set to 5e-4. 
% The learning is set to 5e-4 initially and decays to 0.0 using a cosine annealing schedule. 
% %
% On Tiny-ImageNet, we use SGD optimizer to train the models for 100 epochs with a momentum of 0.9. 
% Weight decay is set to 5e-4. 
% The learning rate ramps up to 5e-4 linearly in the first 2500 iterations and decays to 0.0 using a cosine annealing schedule. 

% 
\smallskip
\boldhdr{Accuracy on CIFAR-10 and CIFAR-100}
% \subsubsection{Results on CIFAR-100}
Table~\ref{tab:cifar_improvement_on_quantization} shows that SQAKD improves the accuracy of EWGS by a large margin for 1, 2, and 4-bit quantization on CIFAR-10 and CIFAR-100. 
Specifically, on CIFAR-10, SQAKD improves EWGS by 0.36\% to 1.28\% on VGG-8 and 0.05\% to 0.39\% on ResNet-20;
and on CIFAR-100, SQAKD improves EWGS by 1.26\% to 3.01\% on VGG-13 and 0.16\% to 1.15\% on ResNet-32.
% As shown in Tables~\ref{tab:cifar_improvement_on_quantization}, SQAKD improves the accuracy of EWGS by 0.36\% to 1.28\% on VGG-8 and 0.05\% to 0.39\% on ResNet-20;
% and on CIFAR-100, SQAKD improves EWGS by 1.26\% to 3.01\% on VGG-13 and 0.16\% to 1.15\% on ResNet-32. 

\smallskip
\boldhdr{Accuracy on Tiny-ImageNet} 
%
% Table~\ref{tab:tiny_imagenet_improvement_on_quantization_large_models} shows the top-1 test accuracy of ResNet and VGG models quantized into 3, 4, and 8 bits on Tiny-ImageNet.
Tables \ref{tab:tiny_imagenet_improvement_on_quantization_large_models} and \ref{tab:tiny_imagenet_improvement_on_quantization_small_models} show that SQAKD consistently improves the accuracy of various quantization methods, including PACT~\cite{choi2018pact}, LSQ~\cite{esser2019learned}, and DoReFa~\cite{zhou2016dorefa}, by a large margin in all cases. 
Specifically, SQAKD improves
1) PACT by 0.41\% to 15.86\%, 
2) LSQ by 0.04\% to 12.03\%, 
and 3) DoReFa 0.3\% to 3.96\% for 3, 4, and 8-bit quantization.

\smallskip
\boldhdr{Convergence Speed} 
Figure~\ref{fig:improvement_on_quantization} illustrate the top-1 test accuracy evolution for 1-bit VGG-18 (CIFAR-10), 4-bit ResNet-32 (CIFAR-100), and 4-bit MobileNet-V2 (Tiny-ImaeNet), respectively, in each epoch during training. 
SQAKD improves the convergence speed of the existing quantization methods on all model architectures.
Furthermore, on MobileNet-V2, SQAKD enables the quantized student to converge much faster than the full-precision teacher whereas the quantization method alone cannot achieve that.
% The results confirm that SQAKD improves EWGS in both converge speed and accuracy.

% Figures~\ref{fig:cifar10_vgg8_improvement_on_quantizatio} and \ref{fig:cifar10_resnet20_improvement_on_quantizatio} illustrate the top-1 test accuracy evolution for VGG-8 and ResNet-20, respectively, in each epoch during training on CIFAR-10. 
% As observed, the convergence speed of the 1-bit VGG-13 and 2-bit ResNet-20 trained under SQAKD that incorporates EWGS is much faster than EWGS alone.
% The results confirm that SQAKD improves EWGS in both converge speed and accuracy.

%\subsection{Transfer Learning}

% \smallskip
% \boldhdr{Inference Speedup} 
% \label{sec:inference_speedup}
% The reduction in model complexity in bit width that SQAKD achieves does translate to real speedup in model inference. 
% We conduct inference experiments using the PyTorch framework and NVIDIA TensorRT on Jetson Nano, a widely used Internet-of-Things (IoT) platform. 
% Table~\ref{tab:inference_speedup} shows our method significantly improves the inference speed by about 3$\times$ for 8-bit quantization using various model architectures, including ResNet-18, MobileNet-V2, ShuffleNet-V2, and SqueezeNet, on Tiny-ImageNet. 
% \input{ablation}
% \input{discussion}
% \input{conclusion}

% \linespread{1.0}

\bibliographystyle{ieeetr}
\bibliography{reference}

\end{document}